# MAT: Motion-Aware Multi-Object Tracking

Shoudong Han,[1*] Piao Huang,[1] Hongwei Wang,[1] En Yu,[1] Donghaisheng Liu,[1] Xiaofeng Pan,[1] Jun Zhao[2]

National Key Laboratory of Science and Technology on Multispectral Information Processing, School of Artificial Intelligence and Automation, Huazhong University of Science and Technology,[1] Nanyang Technological University[2]
{shoudonghan, huangpiao, hongweiwang, yuen, donghaisheng, xiaofengpan}@hust.edu.cn,[1] junzhao@ntu.edu.sg[2]

## Abstract

Modern multi-object tracking (MOT) systems usually model the trajectories by associating per-frame detections. However, when camera motion, fast motion, and occlusion challenges occur, it is difficult to ensure long-range tracking or even the tracklet purity, especially for small objects. Although re-identification is often employed, due to noisy partial-detections, similar appearance, and lack of temporal-spatial constraints, it is not only unreliable and time-consuming, but still cannot address the false negatives for occluded and blurred objects. In this paper, we propose an enhanced MOT paradigm, namely Motion-Aware Tracker (MAT), focusing more on various motion patterns of different objects. The rigid camera motion and nonrigid pedestrian motion are blended compatibly to form the integrated motion localization module. Meanwhile, we introduce the dynamic reconnection context module, which aims to balance the robustness of long-range motion-based reconnection, and includes the cyclic pseudo-observation updating strategy to smoothly fill in the tracking fragments caused by occlusion or blur. Additionally, the 3D integral image module is presented to efficiently cut useless track-detection association connections with temporal-spatial constraints. Extensive experiments on MOT16 and MOT17 challenging benchmarks demonstrate that our MAT approach can achieve the superior performance by a large margin with high efficiency, in contrast to other state-of-the-art trackers.

## 1 Introduction

Multi-object tracking (MOT) plays a crucial role in scene understanding tasks for video analysis. It aims to estimate the trajectories of objects and associate them with per-frame detection results in either online or offline way. With recent progresses on object detection task, tracking-by-detection becomes the preferred paradigm to solve the problem of tracking multiple objects. However, despite the semantic advantages of the dependence on detection, it also turns into a major limitation in complex scenes due to its visible requirements and non-temporal concerns.

Proverbially, the core part of tracking-by-detection paradigm is data association, which is usually performed by using some spatial-scale metrics for the association between consecutive frames, and the re-identification (ReID) is often employed for long-range matching or per-frame association. Some methods like Tracktor++ (Bergmann, Meinhardt, and Leal-Taixé 2019), tending to directly use adjacent-frame detections to carry out spatial-scale association with the assumption of high frame rate and low pedestrian speed. Actually, most tracking scenarios come with a variety of challenges, such as camera motion, fast motion, and occlusion. When that happens, the detector may fail in steadily outputting high-quality detections and large displacement of objects may occur between adjacent frames. So, it is difficult for spatial-scale association to ensure long-range tracking or even tracklet purity in these cases, especially for small objects. To alleviate the spatial-scale inconsistency problem from large displacement or long occlusion, most tracking frameworks adopt a ReID branch to extract objects' embedding features for subsequent feature-level association. For example, DeepSort (Wojke, Bewley, and Paulus 2017) uses a trained ReID model to perform feature extraction, and FairMOT (Zhang et. al. 2020) adds a 512-Dim embedding head after backbone. But ReID-based trackers rely too much on the performance of the detector, the extracted appearance features are often of poor discriminability in complex or low-resolution scenes, such as the partial-detections, noisy pedestrian jamming, blur or similar appearance, and lack of temporal-spatial matching constraints. It is already very time-consuming for ReID computing, the above-mentioned issues may cause a mass of error associations with Identity Switches, and extensive false negatives result from occlusion or blur still cannot be addressed.

In this paper, we abandon the ReID-based association which requires high-quality detections with clean and discriminative appearance. Instead, we focus more on various motion patterns of different objects to realize motion-based prediction and association, and propose an enhanced tracking-by-detection MOT paradigm, namely Motion-Aware Tracker (MAT). Our tracking framework consists of three key modules, they are Integrated Motion Localization (IML), Dynamic Reconnection Context (DRC), and 3D Integral

Image (3DII), respectively. In details, the IML module is designed for the joint prediction of nonrigid pedestrian motion and rigid camera motion, which can be useful for motion-based association. The DRC module determines the motion-based reconnection windows dynamically for different objects, in order to balance the robustness of long-range tracking, based on estimated individual velocity and self-defined intensity of camera motion. Besides, when existing missed detections caused by occlusion or blur, cyclic pseudo-observation updating strategy is included to smooth the predictions of IML nonlinearly as filled tracking fragments. The 3DII module is presented to efficiently cut useless connections among all tracks and detections during data association stage, by imposing the temporal-spatial constraints on each connection. Thus, the tracks within partial-covering region of each detection will be obtained as association candidates integrally in nearly constant time cost, and all the detections will be transferred into multilayered encoding maps.

The contributions of this paper can be summarized as follows:

- We propose an enhanced MOT paradigm, namely Motion-Aware Tracker (MAT), which can be easily extend to any tracking-by-detection architecture with the state-of-the-art performance on MOT16&17 benchmarks.
- We blend the nonrigid pedestrian motion and rigid camera motion seamlessly to balance their compatible issues.
- We design a general dynamic reconnection context module to ensure the robustness and smoother filled tracking fragments for long-range motion-based reconnection.
- We apply the temporal-spatial constraints to filter useless track-detection association connections with lower time cost by 3D integral image encoding.

## 2 Related Work

Recent related studies on tracking-by-detection MOT can be categorized as follows:

**ReID-based data association.** Thanks to the great progresses of object detection over the past few years, most detection-based MOT methods first locate the positions of objects through the detection outputs, then extract the corresponding appearance features via ReID module for long-range matching or per-frame data association. Tracktor++ exploited the bounding box regression of Faster R-CNN (Ren et al. 2015) detector to modify the predicted position of a trajectory in the next frame, and associated the deactivated tracks with detections by a straightforward ReID network. CTracker (Peng et. al. 2020) extended the single-frame regression to the adjacent-frame paired regression, and treated a ReID module as identity-attention to form the detection and tracking into an end-to-end chained structure. CenterTrack (Zhou, Koltun, and Krähenbühl 2020) incorporated CenterNet (Zhou, Wang, and Krähenbühl 2019) as detection module, and used the spatial distance of track-detection center points to calculate the similarity for data association. Furthermore, FairMOT integrated the CenterNet and ReID into one framework for joint spatial-embedding association, which greatly reduced the Identity Switches and maintained the high tracking speed. However, most existing detector-based trackers often fail to model the temporal-spatial relations of large displacement detections when camera motion, fast motion, and occlusion challenges occur. Add the noisy detections and similar appearance, the ReID-based data association may be highly unreliable and the missed detections still cannot be found back. In contrast, our MAT mainly focuses on motion-based prediction, association and reconnection to ensure long-range tracking with high efficiency.

**Motion models for trajectory prediction.** Several trackers resort to motion to predict the temporal-spatial variations of trajectories and compensate the detector failures in complex scenes. The motions in video sequences can be summarized as nonrigid motion (objects like pedestrians) and rigid motion (changing camera pose). The nonrigid motion is commonly described by constant velocity model (Choi and Savarese 2010). In (Yang, Huang, and Nevatia 2011), trajectories were smoothed by observation-based Gaussian distributions. Recently the Kalman Filter tends to be more popular using the provided detections as observations (Bewley et. al. 2016; Wojke et al. 2017). Moreover, the social force models were applied due to complex pedestrian motion in crowded scenarios (Leal-Taixé et. al. 2014). As for the rigid motion caused by camera pose variances, researchers study it mainly in two directions. One is the 3D-information based methods, such as Ego-motion (Wang et. al. 2019) and SFM (Choi and Savarese 2010). The other one is based on affine transformation. Besides, the conditional probability model with recurrent neural network (Fang et. al. 2018) was also proposed to predict the object's position and shape in the next frame. Moreover, the single object tracking (SOT) based methods (Chu and Ling 2019; Chu et al. 2020; Feng et al. 2019; Huang et al. 2020; Y. Xu et al. 2020; Zhu et al. 2018) were gradually adopted to search multiple objects directly. While in these methods, the pedestrian motion and camera motion are always modeled and used separately, and the objects' motion states cannot be estimated precisely. Differently, our MAT designs the Integrated Motion Localization module to blend the pedestrian and camera motions seamlessly to balance their compatible issues, and the camera motion state can be estimated by calculating our defined metric named as intensity of camera motion.

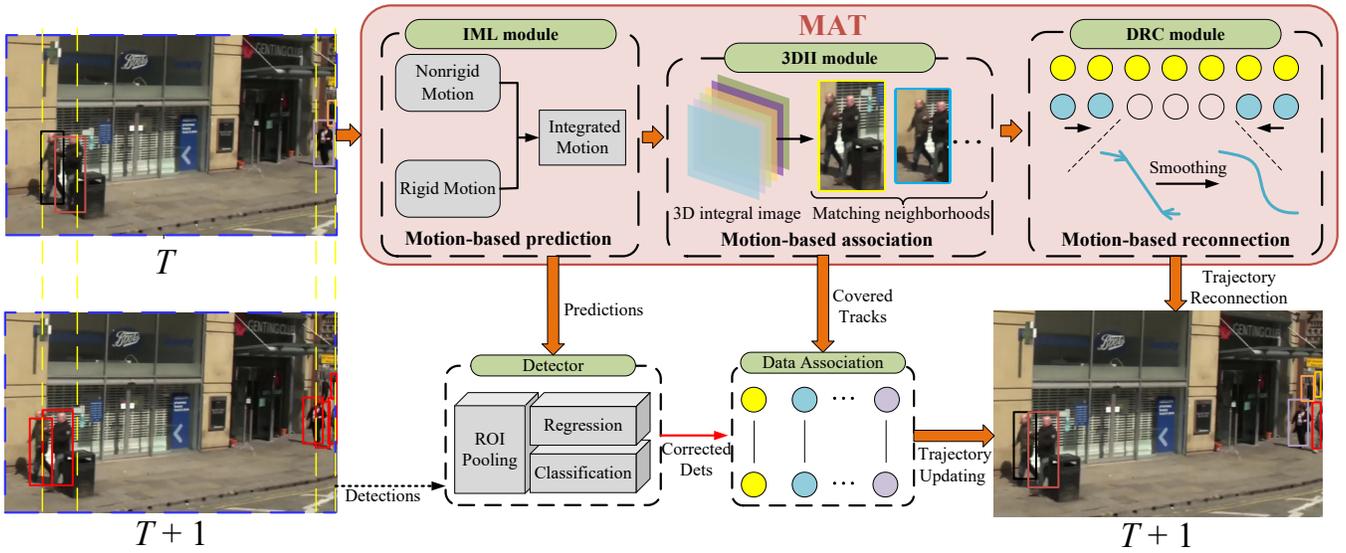

Fig. 1. Illustration of the MAT architecture with a regression-based detector. For a given frame *T*, the IML module is applied to predict each track's position in frame *T*+1 considering both pedestrian and camera motion. Second, all tracks within partial-covering searching region of each corrected detection will be efficiently matched using the 3DII module for motion-based association. Finally, for deactivated trajectories, the DRC module is used to determine their reliable reconnection windows and fill in the tracking fragments smoothly.

## 3 Proposed Method

In this work, we propose an enhanced MOT paradigm MAT, which can be easily included to extend any tracking-by-detection tracker. Here, we employ a regression-based detector like Tracktor++ as our baseline tracker, and treat the MAT predictions and the provided detections as self-defined proposals to obtain corrected detections by passing them into ROI Pooling block for regression and classification. Our MAT architecture is illustrated in Fig. 1, it mainly contains three motion-based modules, namely the Integrated Motion Localization (IML) for track prediction, Dynamic Reconnection Context (DRC) for trajectory reconnection, and 3D Integral Image (3DII) for fast temporal-spatial association.

### 3.1 Integrated Motion Localization (IML)

Given high-quality detections with the simple motion assumption, the data association based on IoU (Intersection-over-Union) metric can obtain admirable tracking results. However, if there exist camera motion, fast motion, occlusion, or low frame rate challenges, the IoU-based trackers without advanced motion compensation may fail in long-range tracking, especially for small objects. Therefore, our IML module focuses on studying in combining rigid camera motion and nonrigid pedestrian motion tightly.

As for camera motion, the pixel alignment among sequential frames can usually be established by the epipolar geometry (Ego) constraints or affine transformations. With the assumption that the objects in adjacent frames have slow motion and constant shapes, then the camera states can be formulated as an optimization problem using Ego motion. Whereas, the fundamental matrix in Ego model needs to be estimated by feature point matching, and it will be seriously interfered by textural pedestrian parts since most of the feature points locate in the areas full of gradient information. In contrast, the global affine transformation is more suitable and robust for approaching the changes of exterior parameters caused by camera motion. Consequently, the Enhanced Correlation Coefficient Maximization (ECC) (Evangelidis and Psarakis 2008) model is included in IML module to estimate the global scene rotation and translation.

Additionally, the Kalman Filter used in DeepSort is our best choice to model the pedestrian motion for high efficiency and flexibility. When blending the camera and pedestrian motions, considering the temporal-spatial consistency and compatibility of different motion patterns, the pedestrian motion is agreed to be processed always before the camera motion, due to the fact that pedestrian motion often plays the principal role in most MOT scenarios. In details, each object's position needs to be predicted by Kalman Filter firstly and then aligned by the ECC model. Thus, our IML model can be simply established as below.

$$\begin{cases} s_{t+1} = warp(Fs_t) \\ P_{t+1} = FP_tF^T + Q \end{cases} \quad (1)$$

where $F, Q, P, s$ respectively denote the state transition matrix, process uncertainty, covariance matrix, and state estimate of Kalman Filter. Here $s = [Box, V_{Box}]^T$, $Box$ represents the predicted coordinates of target bounding box, and $V_{Box}$ contains the estimated velocities of all elements in $Box$. Besides, the $warp$ operation denotes the ECC alignment, notice that it is only applied to the $Box$ part of $Fs_t$.

## 3.2 Dynamic Reconnection Context (DRC)

When heavy occlusion or blur happens, it surely will lead to plenty of missed detections and force the trajectories to be deactivated temporally. To guarantee the robustness of long-range tracking, this section proposes the DRC module for motion-based reconnection, based on the estimated motion states and predictions of above IML module. It includes two parts, which are dynamic motion-based reconnection mechanism, and cyclic pseudo-observation trajectory filling.

**Dynamic Motion-Based Reconnection Mechanism**

In crowded multi-object scenarios, there are lots of interactions and occlusions among different pedestrians, which will cause trajectories to break or overlap.

The normal solution is to use ReID embedding metric to reconnect deactivated trajectories. But this way cannot work well for discriminating small scale, occluded, blurred, and similar objects. Moreover, ReID-based association is a fully connected matching strategy, which is time-consuming and unreliable due to the absence of temporal-spatial constraints. Worst of all, it still cannot recover any pieces of the tracking fragments even if successfully reconnected.

To alleviate the problems mentioned above, we design the dynamic motion-based reconnection mechanism. In details, we suggest to enable the full-lifecycle predictions for all trajectories using our IML module, even though the trajectory is deactivated temporally due to association failure, which can be employed as the temporal-spatial cues to wait for the following track-detection association.

In addition, on account of various motion patterns of different objects, such as the diverse moving speed of pedestrian and scene, it is unsuitable to set a unified reconnection window for all deactivated trajectories. Although our IML module can work steadily with continuous observation updating, the pure inertia prediction without observation updating during deactivated period may become more and more unreliable especially for trajectories with fast pedestrian and camera motion. Therefore, so as to balance the robustness of long-range tracking, we propose an adaptive mechanism to determine the motion-based reconnection window dynamically for each object based on its current motion states as shown in formula (2).

$$L_{rec} = L_{max} * e^{-[\alpha * I_{Cam} + (1-\alpha) * |V_{Box}|]} \quad (2)$$

where $L_{rec}$ denotes the dynamic length of patient reconnection window, $L_{max}$ represents the maximal length of possible reconnection, $|V_{Box}|$ denotes the estimated individual Kalman velocity as defined in the description of formula (1), and $\alpha$ is an adjustment coefficient describing the weights of different motion patterns. Besides, $I_{Cam}$ is the intensity of camera motion that can be defined as formula (3).

$$I_{Cam} = 1 - \frac{W \times R}{\|W\|_2 \times \|R\|_2}, R = [I; O] \quad (3)$$

Here $W$ denotes the vectorization of affine matrix in ECC model, and $R$ means the affine matrix of static frames. $I$ is the identity matrix and $O$ is the all-zero matrix.

**Cyclic Pseudo-Observation Trajectory Filling**

With the help of dynamic motion-based reconnection, most deactivated trajectories can be reconnected successfully in longer windows with highly reliable temporal-spatial constraints. Although the predictions of IML module during the deactivated period can be directly retained to construct the missed tracking fragment after reconnection, they may still be far away from the true trajectory, and not smooth enough with a saltation at reconnection point, as shown in Fig. 2.

Therefore, we include the cyclic pseudo-observation trajectory filling strategy to smooth the predictions of IML nonlinearly as the final filled tracking fragments. Assuming that some trajectory is interrupted from active to deactivated at point **A**, and reconnected successfully at point **B** thought track-detection association as demonstrated in Fig. 2. Then our trajectory filling strategy can be described as the following three steps. 1) Linear initialization, using the linear interpolation algorithm to generate the initial boxes for all frames between point **A** and **B** with uniform changes in position and scale. 2) Forward IML updating, using the initial boxes obtained at last step as the pseudo-observations to forward update the trained IML model at point **A** frame by frame until to point **B**. 3) Backward IML updating, training a new IML model backward with several frames of the tracklet after point **B**, and use the predictions obtained at last step as the pseudo-observations to backward update the trained IML model at point **B** frame by frame until to point **A**. After these three steps, the final corrected predictions are employed to fill in the tracking fragment.

Consequently, the final filled tracking fragment will take the linear initialization as the baseline, and the first half is mainly forward smoothed by the IML model at point **A**, while the latter half is mainly backward smoothed by the IML model at point **B**. This kind of cyclic nonlinear smoothing can further reduce the offset between the filled trajectory and true trajectory.

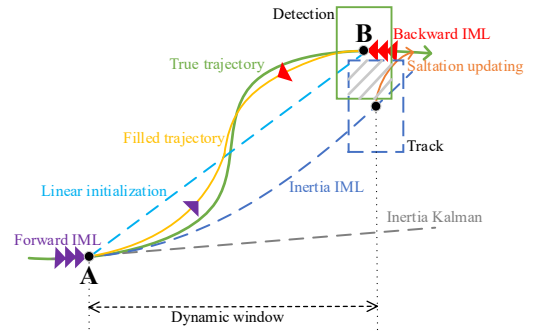

Fig. 2. Illustration of the cyclic pseudo-observation trajectory filling process.

## 3.3 3D Integral Image (3DII)

In a dense multi-object scenario, the object number can be dozens or even hundreds, which imposes a significant time cost on the calculation of cost matrix among predicted tracks and corrected detections in fully connected manner. Hence, during data association stage, we present the 3DII module to efficiently cut useless track-detection connections by transferring all detections into the multilayered encoding maps.

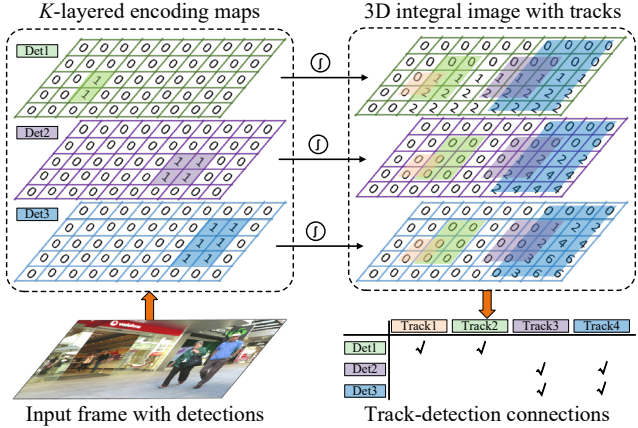

Fig. 3. The encoding and filtering process of 3DII module.

As shown in Fig. 3, the input frame with $K$ detections is divided into $M \times N$ cells, then the $K$-layered encoding maps will be constrained with size of $M \times N$ for each layer. Each detection corresponds to one layer, the map for this layer is initialized with all zeros, and the cells within partial-covering region of this detection bounding box will be encoded as ones. Therefore, the 3D integral image of the $K$-layered encoding maps at location $(m,n)$ can be modeled as the sum of one-valued cells from $(0, 0)$ to $(m, n)$ for each layer as:

$$I_{3D}(m,n) = \sum_{m' \leq m, n' \leq n} f(m', n') \quad (4)$$

where $I_{3D}$ represents the $K$-dim integral image, and $f$ denotes the $K$-layered encoding maps with binary values.

To accelerate the calculation of formula (4), we can simplify its computing process using dynamic programming.

$$I_{3D}(m,n) = I_{3D}(m, n-1) + I_{3D}(m-1, n) \\ - I_{3D}(m-1, n-1) + f(m,n) \quad (5)$$

For each coming track, firstly, the diagonal coordinates of its bounding box can be mapped to the coordinate system of $f$ as $[x_1, x_2, y_1, y_2]$. Using the above 3D integral image, the detections within partial-covering region of this track will be obtained directly in nearly constant time cost.

$$I_{3D}(x_1{:}x_2, y_1{:}y_2) = I_{3D}(x_2, y_2) + I_{3D}(x_1 - 1, y_1 - 1) \\ - I_{3D}(x_1 - 1, y_2) - I_{3D}(x_2, y_1 - 1) \quad (6)$$

where the non-zero value in computed $I_{3D}(x_1{:}x_2, y_1{:}y_2)$ denotes the found nearby detection.

Notice that most of the operations in 3DII are assignment, addition and subtraction, so the time cost for filtering useless track-detection connections can be dramatically reduced. Besides, the 3DII is helpful to enable the temporal-spatial constraints at any scale level by just changing the cell size.

## 4 Experiments

### 4.1 Experiment Setup

Experiments are conducted on two widely used challenging MOT benchmarks: MOT16 and MOT17 (Anton et al. 2016). Although both of them contain the same 7 training sequences and 7 test sequences, they provide the different public detections and different ground truth labels. Such as, MOT17 includes DPM (Felzenszwalb et al. 2010), Faster R-CNN, and SDP (Yang, Choi, and Lin 2016) with increasing performance, but MOT16 only includes the DPM detector. These benchmarks consist of extensive challenging pedestrian tracking and detection with frequent occlusion, and the scenes are heavily crowed and vary in the camera poses, object scales and frame rates.

**Evaluation Metrics.** Evaluation is carried out according to the widely accepted CLEAR MOT metrics (Yang et al. 2016), including the Multiple Object Tracking Accuracy (MOTA), ID F1 Score (IDF1), False Positives (FP), False Negatives (FN), Identity Switches (IDS), and Tracker Speed (Hz), et al. Among these performance metrics, the MOTA and IDS can quantify the main two aspects as trajectory accuracy and purity.

**Implementation Details.** All the experiments are implemented using PyTorch and run on a desktop with a CPU of 10 cores@2.2GHz and a RTX2080Ti GPU. As for the detector, we re-implement Cascaded RCNN (Cai and Vasconcelos 2018) by changing the anchor scales to {32, 64, 128, 256, 512} and anchor aspects to {1.0, 2.0, 3.0} with pre-trained network parameters on COCO datasets (Chen et al. 2014). The detector is simply trained on the training dataset of MOT17 by taking the samples with label 1 and visibility above 0.1 as positives, and the others as negatives. Random flipping is adopted to realize the data augmentation, and all the clip operations for ground truth boxes, anchors, and proposals are canceled to obtain complete detections even beyond the boundaries. The detector training process takes 30000 iterations with the batch size of 8, using an initial learning rate of $e^{-8}$ and the multi-step SGD optimizer. The naive NMS (Dalal and Triggs 2005) is employed with the confidence threshold of 0.05 and IoU threshold of 0.5. Considering the MAT settings, the simple IoU metric is used to calculate the cost matrix of Kuhn-Munkres (KM) algorithm (Kuhn 2010) for motion-based association, the Kalman Filter is simplified as a constant velocity model (J. Peng et al. 2020), $L_{max}$ and $\alpha$ in formula (2) are fixed as 120 and

0.95, the 3DII map size is set as 16×8, and the trajectories whose length less than 5 frames are removed for post-processing.

### 4.2 Ablation Study

Our ablation study mainly aims to reveal the performance of our motion-aware designs, which don't depend on the training of ground truth. Therefore, to avoid the possible interference caused by the quality of detector, we directly compare and evaluate the following implementations on MOT17 training dataset for convenience.

| Method | MOTA↑ | IDF1↑ | FP↓ | FN↓ | IDS↓ |
|---|---|---|---|---|---|
| Baseline | 68.0 | 69.2 | 3749 | 102709 | 1407 |
| Baseline+Ego | 45.5 | 41.4 | 14983 | 158850 | 9937 |
| Baseline+ECC | 68.2 | 69.1 | 3949 | 102094 | 1083 |
| Baseline+IML | 68.7 | 71.1 | **3430** | 101017 | 850 |
| Baseline+IML+DW | 71.1 | 74.6 | 10422 | 86360 | **682** |
| **Baseline+IML+DRC** | **71.5** | **74.8** | 9711 | **85653** | 701 |

Table 1. Ablation study in terms of different motion models and motion-based reconnection mechanisms.

As shown in Table 1, the Baseline represents our re-implemented Detector with Kalman, without any guidance of camera motion and dynamic reconnection, which is the simplest version of our MAT. Notice that the Baseline still includes a simple reconnection mechanism like Tracktor++, where the deactivated trajectories can be reconnected within a fixed window of 10 frames and filled using inertia predictions. The Baseline+ means to replace some module in Baseline, such as the motion or reconnection, with other choices. For example, Baseline+Ego, Baseline+ECC, Baseline+IML replace the Kalman motion with Ego, ECC, and our IML respectively, Baseline+IML+DW just replace the fixed window of Baseline+IML with our dynamic-window (DW) referring to formula (2), and Baseline+IML+DRC is the non-accelerated implementation of MAT without 3DII module.

The results in Table 1 can verify that:

1) The Ego model cannot handle the motion alignments well just like the analysis given in Section 3.1, the ECC model can achieve nearly the same performance as Kalman Filter with much lower IDS, and our IML model can bring the Baseline tracker an overall promotion on all metrics especially the IDS, which tells the powerful motion pattern compatibility of the IML module.

2) Based on our IML module, it is also demonstrated that the further employment of our DW or complete DRC reconnection mechanism, can significantly improve the tracking performance with great margins on the MOTA, IDF1, FN and IDS but little sacrifice of FP. In other words, for motion-based reconnection, it is essential to design an adaptive individual window considering the various motion patterns. Besides, our proposed cyclic pseudo-observation trajectory filling strategy (Baseline+IML+DRC) is proved to be more robust than using inertia predictions (Baseline+IML+DW), with lower FP, FN and higher MOTA, IDF1. Note that the IDS of DW is slightly smaller than DRC since much more long-range trajectories are tracked by DRC.

**Dynamic vs. fixed window when both using our cyclic pseudo-observation trajectory filling**. The above ablation experiments (Baseline+IML+DW vs. Baseline+IML) have confirmed that our design of dynamic reconnection window can obtain great improvement over fixed one even using inertia predictions to fill in the tracking fragments. As shown in Fig. 4, this part gives more details about dynamic vs. fixed window when using our powerful cyclic pseudo-observation trajectory filling under the same maximal length of possible reconnection $L_{max}$. The curves in Fig. 4 reveal that with the increase of $L_{max}$, the dynamic-window based DRC module gradually outperforms the one based on fixed window with a higher-level of MOTA, and the peak point is up to 120 frames to enable the long-range tracking tasks.

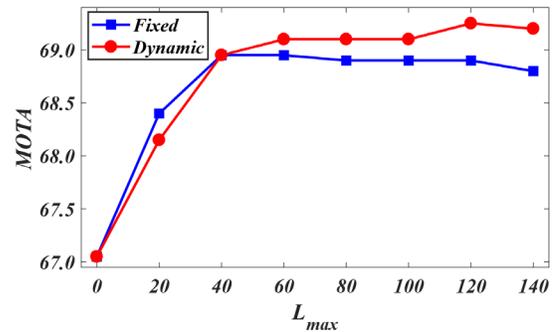

Fig. 4. Dynamic vs. fixed window motion-based reconnection under the different maximal lengths.

**3DII module**. During the data association stage, it is necessary to cut useless track-detection connections to impose more temporal-spatial constraints. If a detection has some overlap with the extended region of a track's bounding box, then the detection will be assigned as a candidate of the track. To verify the acceleration performance of our 3DII module to finish this job, we also present an IoU-based fully-connected method to filter the connections without any intersection, named as IoU-filter for comparison. The speed comparison is shown in Fig. 5, where the numbers of tracks and detections are simplified as the same when applying track-detection association. The larger track-detection number denotes the denser multi-object scenario. The results in Fig. 5 demonstrate the significant advantage of our 3DII module in acceleration with nearly constant time cost, over the IoU-filter, especially when there exist a large number of pedestrians per frame. Note that our 3DII module can easily enable the temporal-spatial constraints at any extended scale level by just changing the cell size, but the IoU-filter can only judge if there is an intersection or not.

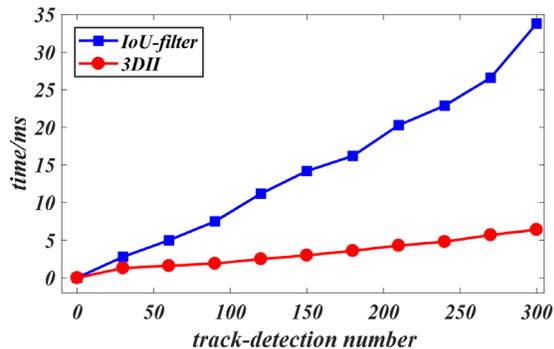

Fig. 5. Speed comparison using our 3DII module or IoU-filter.

### 4.3 Evaluation on Benchmarks

The performance of our MAT approach has been evaluated on both the MOT16 and MOT17 test datasets. We also compare our tracker with other best public and private trackers, which have been officially published and peer reviewed in these benchmarks. The comparisons are shown in the Table 2, where both the state-of-the-art offline and online MOT methods are included, such as the offline trackers MPNTracker (Brasó and Leal-Taixé 2020) and Lif_T (Hornakova et. al. 2020), and online trackers (denoted using **O**) like Trackor++, DeepMOT (Y. H. Xu et al. 2020), UMA (Yin et. al. 2020), UnsupTrack (Karthik, Prabhu, and Gandhi 2020), CenterTrack, JDE (Wang et al. 2020), FairMOT, CTracker, and Tube_TK (Pang et. al. 2020).

As demonstrated in Table 2, our tracker MAT significantly outperforms all these existing state-of-the-art MOT methods on MOT16 and MOT17 challenging benchmarks by a large margin, especially in terms of the MOTA and FN. Besides, the metrics of IDF1 and IDS are almost the best for our MAT, no matter using the public or private detector. Also, our MAT has much lower time cost, as shown in the Hz column, compared to most of the online or offline trackers. It should be noted that, the main concern of this paper is motion-aware designs but not the optimization of detector, so the performance of our re-implemented detector may not be as good as other compared private detectors in terms of accuracy and speed. Even so, our MAT still wins the championship on the primary metric MOTA, which reflects the overall tracking performance.

To summarize, our proposed MAT can achieve the superior tracing performance in MOTA and IDF1 mainly due to the powerful motion pattern compatibility of the IML module, which guarantees the continuity and purity of trajectories. Additionally, by further using our proposed DRC module, the long-range motion-based tracking and reconnection come true with a dramatical decreasing of FN and IDS. Furthermore, in theory our motion-aware tracker can obtain super high tracking speed given a fast detector, and our 3DII module can address another speed bottleneck for data association, imposing more temporal-spatial constraints and filtering useless track-detection connections with nearly constant time cost.

| Method | MOTA↑ | IDF1↑ | FP↓ | FN↓ | IDS↓ | Hz↑ |
|---|---|---|---|---|---|---|
| Public Detection on MOT16 | | | | | | |
| MPNTracker | 58.6 | 61.7 | 4949 | 70252 | **354** | 6.5 |
| Lif_T | 61.3 | 64.7 | 4844 | 65401 | 389 | 0.5 |
| Trackor++(**O**) | 54.4 | 52.5 | 3280 | 79149 | 682 | 2.0 |
| DeepMOT(**O**) | 54.8 | 53.4 | **2955** | 78765 | 645 | - |
| UMA(**O**) | 50.5 | 52.8 | 7587 | 81924 | 685 | 5.0 |
| UnsupTrack(**O**) | 62.4 | 58.5 | 5909 | 61981 | 588 | 1.9 |
| **Ours(O)** | **67.7** | **69.6** | 6337 | **52234** | 379 | **11.5** |
| Private Detection on MOT16 | | | | | | |
| JDE(**O**) | 64.4 | 55.8 | - | - | 1544 | 18.8 |
| FairMOT(**O**) | 68.7 | **70.4** | - | - | 953 | 25.9 |
| CTracker(**O**) | 67.6 | 57.2 | **8934** | 48305 | 1897 | **34.4** |
| Tube_TK(**O**) | 66.9 | 62.2 | 11544 | 47502 | 1236 | 1.0 |
| **Ours(O)** | **70.5** | 63.8 | 11318 | **41592** | **928** | 9.1 |
| Public Detection on MOT17 | | | | | | |
| MPNTracker | 58.8 | 61.7 | 17413 | 213594 | **1185** | 6.5 |
| Lif_T | 60.5 | 65.6 | 14966 | 206619 | 1189 | 0.5 |
| Trackor++(**O**) | 53.5 | 52.3 | 12201 | 248047 | 2072 | 2.0 |
| CenterTrack(**O**) | 61.4 | 53.3 | 15520 | 196886 | 5326 | - |
| DeepMOT(**O**) | 53.7 | 53.8 | **11731** | 247447 | 1947 | - |
| UMA(**O**) | 53.1 | 54.4 | 22893 | 239534 | 2251 | 5.0 |
| UnsupTrack(**O**) | 61.7 | 58.1 | 16872 | 197632 | 1864 | |
| **Ours(O)** | **67.1** | **69.2** | 22756 | **161547** | 1279 | **11.5** |
| Private Detection on MOT17 | | | | | | |
| FairMOT(**O**) | 67.5 | **69.8** | - | - | 2868 | 25.9 |
| CTracker(**O**) | 66.6 | 57.4 | **22284** | 160491 | 5529 | **34.4** |
| CenterTrack(**O**) | 67.3 | 59.9 | 23031 | 158676 | 2898 | - |
| Tube_TK(**O**) | 63.0 | 58.6 | 27060 | 177483 | 4137 | 3.0 |
| **Ours(O)** | **69.5** | 63.1 | 30660 | **138741** | 2844 | 9.0 |

Table 2. Comparisons of state-of-the-art MOT methods on MOT16 and MOT17 test datasets.

## 5 Conclusion

In this paper, we propose an enhanced multi-object tracking paradigm named as Motion-Aware Tracker (MAT), which mainly focus on the aware designs of motion-based prediction, reconnection, and association. It mainly contains three modules, namely the Integrated Motion Localization (IML) for blending various motion patterns compatibly, Dynamic Reconnection Context (DRC) for dynamically window determining and smoothly trajectory filling, and 3D Integral Image (3DII) for fast temporal-spatial filtering of useless track-detection connections. Extensive experiments are conducted on the widely used MOT16&17 challenging benchmarks, and all the results demonstrate the superiority of our MAT approach in terms of the state-of-the-art accuracy and high efficiency. Besides, the architecture of our MAT is very simple and general, which can be easily embedded to extend any tracking-by-detection tracker or video detector.

## Acknowledgment

This work was supported by the National Natural Science Foundation of China under Grant No. 61105006; Open Fund of Key Laboratory of Image Processing and Intelligent Control (Huazhong University of Science and Technology), Ministry of Education under Grant No. IPIC2019-01; and the China Scholarship Council under Grant No. 201906165066.

# MAT: Motion-Aware Multi-Object Tracking

## (Supplementary Material)

**Shoudong Han,[1]* Piao Huang,[1] Hongwei Wang,[1] En Yu,[1] Donghaisheng Liu,[1] Xiaofeng Pan,[1] Jun Zhao[2]**

National Key Laboratory of Science and Technology on Multispectral Information Processing, School of Artificial Intelligence and Automation, Huazhong University of Science and Technology,[1] Nanyang Technological University[2]
{shoudonghan, huangpiao, hongweiwang, yuen, donghaisheng, xiaofengpan}@hust.edu.cn,[1] junzhao@ntu.edu.sg[2]

### Abstract

To facilitate the reproduction and further reveal the superiority of our Motion-Aware Tracker (MAT) approach, this supplementary material adds its structured pseudocode with public or private detector first. Besides, more implementation details of our re-implemented detector in terms of architecture and training are included. In addition, we provide some additional sensitivity analysis experiments about the alpha in formula (2) and backward tracklet length in Fig. 2 of our main work. Moreover, we complement some detailed experimental results and analysis for each sequence with some typical qualitative results, and the visualization of trajectory filling using our DRC module.

## 1 Detailed Implementation

In order to successfully reproduce the experimental results of our main work, here we further detail the implementation of MAT and re-implemented detector.

### 1.1 Algorithm Representation of MAT

As shown in Algorithm 1, we merge the processing logic of our MAT approach with public or private detector, and provide the corresponding pseudocode representation of Fig. 1 and Section 3 of our main work.

Notice that, as illustrated in line 9, we employ the simplest IoU metric to evaluate the temporal-spatial similarity of candidate track-detection connection. Moreover, based on these filtered connections using 3DII module, we can further kill some unreliable connections with low IoU, and the IoU threshold of valid candidate connection can be set as 0.3 before carrying out the KM algorithm as line 10.

### 1.2 Details of Re-implemented Detector

Our re-implemented Cascaded RCNN (Cai and Vasconcelos 2018) detector is built upon the famous Detectron2 framework (Wu et al. 2019) provided by Facebook AI Research, which implements a large set of state-of-the-art object detection algorithms.

**Input:** Video sequence $I = \{I_1, I_2, \cdots, I_S\}$, $public$ or $private$ detections $\mathcal{D} = \{D_1, D_2, \cdots, D_S\}$.
**Output:** Trajectories $\mathcal{T}$.

1. Initialize trajectories with null that $\mathcal{T} \leftarrow \phi$;
2. $\mathcal{L}_k$: Deactivated length of $\mathcal{T}_k$, initialized by 0;
3. **for** $t = 1, \cdots, S$ **do**
4.     $T_t \leftarrow$ Obtain tracks by applying IML to $\mathcal{T}$ referring to formula (1);
5.     **if** $private$ **then**
6.         $B_t \leftarrow D_t$;
7.     **else**
8.         $B_t \leftarrow NMS(Regress\_and\_Classify(\{D_t, T_t\}))$;
9.     $Cost \leftarrow$ Calculate cost matrix of $3DII(B_t, T_t)$ by IoU metric;
10.    Associate the $B_t$ with $T_t$ using KM algorithm according to $Cost$;
11.    **for** $\mathcal{T}_k \in \mathcal{T}$ **do**
12.        Update the $L_{rec}$ of $\mathcal{T}_k$ referring to formula (2);
13.        **if** $T_{t,k}$ is associated with $B_{t,i}$ **then**
14.            $\mathcal{T}_k \leftarrow \mathcal{T}_k + B_{t,i}$;
15.            Fill in the tracking fragment of $\mathcal{T}_k$ using DRC;
16.            $\mathcal{L}_k = 0$;
17.        **else**
18.            **if** $\mathcal{L}_k > L_{rec}$ **then**
19.                Stop further tracking of $\mathcal{T}_k$;
20.            **else**
21.                Hold $T_{t,k}$ with $\mathcal{T}_k$ temporarily for further IML;
22.                $\mathcal{L}_k + +$;
23.    **for** $B_{t,j} \in B_t$ **do**
24.        **if** $B_{t,j}$ is not associated with anyone of $T_t$ **then**
25.            $\mathcal{T} \leftarrow \mathcal{T} + B_{t,j}$;
26. Remove ultra-short trajectories from $\mathcal{T}$ for post-processing.

Algorithm 1. Tracking process of our MAT approach using public or private detector.

In details, we choose ResNet50 (He et al. 2016) and FPN (Lin et al. 2017) as the backbone and neck respectively for feature extraction, and the output features of P2-P6 are used to generate proposals by RPN (Ren et al. 2015). The prediction head consists of three identical and cascaded RCNN (Ren et al. 2015) heads, each RCNN head includes a

RoIAlign (He et al. 2018) module for feature pooling, and a bounding box regression branch and a classification branch. For the first two RCNN heads, the output of bounding box regression branch is used as the input of RoIAlign of the next RCNN head. Finally, the output of the third RCNN head is passed into the naive NMS algorithm to obtain the final object detection results. In addition, the classification branch uses the cross-entropy loss, the regression branch employs the smooth L1 loss, and the total loss function is the sum of the losses of these three RCNN heads.

Most training settings refer to the default manners as mentioned in Cascaded RCNN and our main work. Except that, the first 2000 iterations are used for model warm-up, and after 15000 and 25000 iterations, the learning rate decreases to the 0.1 and 0.01 of the initial value respectively.

## 2 More Experiments

For the sake of further demonstrating superior performance of our MAT approach, here we add more detailed experiments about parameter sensitivity, each sequence, qualitative comparison, and visualization.

### 2.1 Parameter Sensitivity Analysis

According to the experiments in Fig. 4 of our main work, we demonstrate that the maximal possible reconnection length $L_{max}$ in formula (2) can be up to 120 frames for better long-range tracking performance. Here we fix it as 120 and then test the performance sensitivity of another parameter $\alpha$ in formula (2), as shown in Fig. 1. Notice that the other experimental settings of this section are the same as the ablation study of our main work. The results in Fig. 1 reveal that, for better MOTA performance, the intensity of camera motion should play the main role in determining the dynamic length of patient reconnection window when compared with the estimated individual Kalman velocity. Nevertheless, directly remove $|V_{Box}|$ from the formula (2) and set $\alpha$ as 1 is not the best option, and our final MOTA performance is not very sensitive with the variation of parameter $\alpha$. Therefore, as illustrated in our main work, $\alpha$ in formula (2) can be fixed as 0.95 for all the experiments.

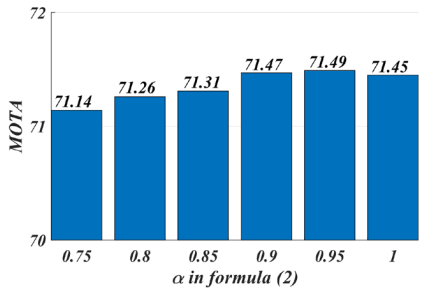

Fig. 1. Sensitivity analysis experiment of $\alpha$ in formula (2) of our main work.

In addition, when discussing the cyclic pseudo-observation trajectory filling process in Fig. 2 and Section 3.2 of our main work, there is a parameter in the third step for backward IML updating, named as backward tracklet length. The backward tracklet consists of several tracks after point **B**, which is used for training a backward IML model. Here we test the performance sensitivity of this parameter, the corresponding MOTA results of different backward tracklet length are shown in Fig. 2. It is verified that our final MOTA performance is almost completely insensitive with the backward tracklet length. Therefore, according to the results of Fig. 2, we fix this parameter as 3 for all the experiments.

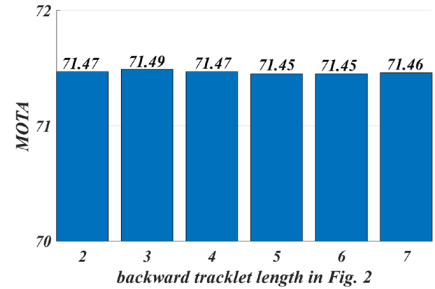

Fig. 2. Sensitivity analysis experiment of backward tracklet length in Fig. 2 of our main work.

### 2.2 Detailed Experimental Results

The detailed experimental results of our MAT approach for each sequence on MOT16 and MOT17 test datasets, corresponding to the results of Table 2 in our main work, are displayed in Table 1 and Table 2 here. These detailed sequence results can also be found in the official MOTChallenge web page available at https://motchallenge.net/.

| Sequence | MOTA↑ | IDF1↑ | FP↓ | FN↓ | IDS↓ |
|---|---|---|---|---|---|
| Public Detection on MOT16 | | | | | |
| 01-DPM | 51.6 | 56.8 | 92 | 2,995 | 8 |
| 03-DPM | 79.2 | 77.4 | 2,444 | 19,179 | 75 |
| 06-DPM | 56.1 | 61.8 | 1,173 | 3,818 | 74 |
| 07-DPM | 57.8 | 56.4 | 800 | 6,034 | 60 |
| 08-DPM | 45.9 | 49.7 | 454 | 8,547 | 56 |
| 12-DPM | 60.7 | 68.6 | 352 | 2,888 | 21 |
| 14-DPM | 46.5 | 57.9 | 1,022 | 8,773 | 85 |
| Private Detection on MOT16 | | | | | |
| 01-Ours | 50.4 | 44.9 | 339 | 2793 | 43 |
| 03-Ours | 86.0 | 72.5 | 2556 | 11,828 | 244 |
| 06-Ours | 62.1 | 65.2 | 1049 | 3,204 | 123 |
| 07-Ours | 54.9 | 47.6 | 1,646 | 5588 | 134 |
| 08-Ours | 43.0 | 40.1 | 389 | 9038 | 114 |
| 12-Ours | 60.0 | 62.6 | 440 | 2834 | 44 |
| 14-Ours | 38.1 | 49.5 | 4,899 | 6,307 | 226 |

Table 1. Detailed sequence results of our MAT approach on MOT16 test dataset corresponding to the Table 2 of our main work.

| Sequence | MOTA↑ | IDF1↑ | FP↓ | FN↓ | IDS↓ |
|---|---|---|---|---|---|
| Public Detection on MOT17 | | | | | |
| 01-DPM | 51.6 | 56.8 | 83 | 3,031 | 8 |
| 03-DPM | 80.1 | 77.8 | 1,976 | 18,825 | 75 |
| 06-DPM | 56.2 | 61.4 | 1,101 | 3,992 | 73 |
| 07-DPM | 55.4 | 54.9 | 832 | 6,625 | 71 |
| 08-DPM | 37.0 | 42.8 | 385 | 12,865 | 57 |
| 12-DPM | 58.7 | 67.2 | 327 | 3,235 | 21 |
| 14-DPM | 46.5 | 57.9 | 1,022 | 8,773 | 85 |
| 01-FRCNN | 48.0 | 59.5 | 564 | 2,781 | 10 |
| 03-FRCNN | 80.1 | 78.9 | 1,765 | 19,021 | 67 |
| 06-FRCNN | 58.8 | 62.7 | 1,363 | 3,407 | 85 |
| 07-FRCNN | 55.8 | 57.5 | 893 | 6,509 | 66 |
| 08-FRCNN | 35.3 | 43.5 | 461 | 13,145 | 54 |
| 12-FRCNN | 52.5 | 61.7 | 451 | 3,649 | 19 |
| 14-FRCNN | 42.6 | 55.4 | 2,479 | 8022 | 103 |
| 01-SDP | 53.6 | 56.2 | 369 | 2,612 | 15 |
| 03-SDP | 85.4 | 79.4 | 2,504 | 12,666 | 98 |
| 06-SDP | 56.9 | 61.8 | 1,748 | 3,227 | 101 |
| 07-SDP | 57.5 | 58.1 | 1,045 | 6,058 | 74 |
| 08-SDP | 37.4 | 43.8 | 432 | 12,745 | 56 |
| 12-SDP | 58.0 | 66.0 | 448 | 3,168 | 24 |
| 14-SDP | 46.9 | 56.3 | 2,508 | 7,191 | 117 |
| Private Detection on MOT17 | | | | | |
| 01-Ours | 50.0 | 44.8 | 340 | 2839 | 43 |
| 03-Ours | 87.0 | 72.8 | 1997 | 11388 | 245 |
| 06-Ours | 63.9 | 65.3 | 860 | 3265 | 127 |
| 07-Ours | 54.8 | 46.8 | 1492 | 6005 | 146 |
| 08-Ours | 35.6 | 35.0 | 229 | 13266 | 117 |
| 12-Ours | 58.1 | 61.3 | 411 | 3177 | 44 |
| 14-Ours | 38.2 | 49.5 | 4891 | 6307 | 226 |

Table 2. Detailed sequence results of our MAT approach on MOT17 test dataset corresponding to the Table 2 of our main work.

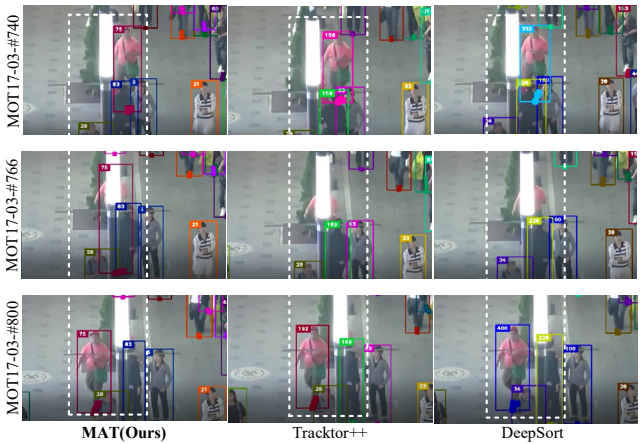

Fig. 3. Qualitative comparison of our MAT with Tracktor++ and DeepSort in a typical occlusion scenario. The white dotted rectangle indicates the tracking differences of the red target pedestrian.

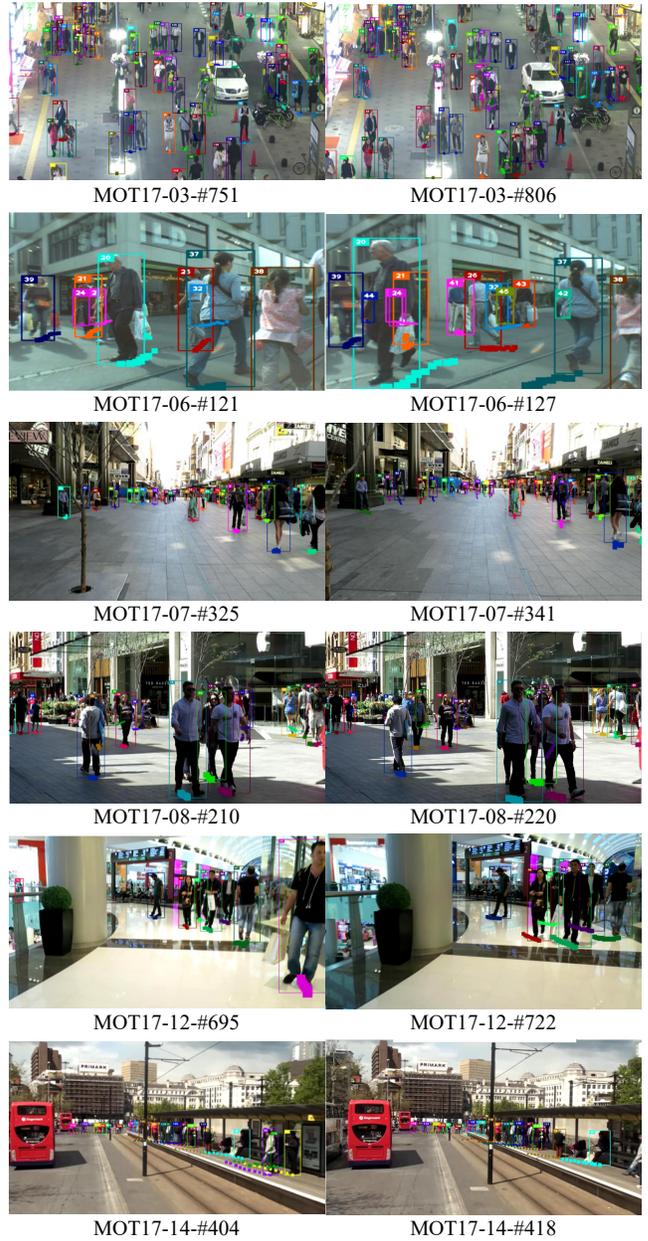

Fig. 4. Visualization of tracking results of our MAT approach using SDP detector on MOT17 test dataset.

The detailed sequence results in Table 1 and Table 2 suggest the same conclusions as our main work. Our main innovations are the motion-aware designs but not the optimization of detector, so the sequence results of our re-implemented detector may not be good enough for all sequences. Even so, since sequences 03 and 06 include extensive occluded pedestrians with fast or camera motion, our MAT can work better with the help of proposed IML and DRC modules. Consequently, the detailed results in Table 1 and Table 2 also verify the effectiveness and motion-aware ability of our MAT design from the side.

Moreover, as displayed in Fig. 3, we select a typical occlusion scenario to qualitatively compare our MAT with two most related trackers, namely the Tracktor++ and DeepSort. The tracking results of our MAT and Tracktor++ are based on the public SDP detector, and the DeepSort uses its private detector. The red target pedestrian in the white dotted rectangle is occluded when walking through the streetlight. It can be seen from Fig. 3 that Tracktor++ and DeepSort are both unable to accurately track this man, and treat this target pedestrian as two different identities before and after being occluded, which will obviously cause False Negatives and Identity Switches. In contrast, our MAT can solve this problem effectively through the proposed IML and DRC modules, and maintain the trajectory accuracy and purity of the occluded object.

Other detailed qualitative results are shown in Fig. 4. All of the test sequences are tracked by our MAT with SDP detections as observations. Intuitively, our proposed MAT can obtain precise tracking boxes of pedestrians with consistent identities. Also, our MAT is robust enough to the irregular camera motions (such as the MOT17-06 and MOT17-14), crowded scenes (such as MOT17-03), and different camera viewpoints (such as MOT17-03 and MOT17-07). Especially on the MOT17-14 sequence, which is captured by a fast-moving camera that is mounted on bus in a busy intersection. Our proposed MAT can still be able to track all the pedestrians in a stable and persistent way.

### 2.3 Visualization of Trajectory Filling

The above supplementary experiments have reflected the overall tracking performance of our MAT approach quantitatively and qualitatively. Here, to further highlight the motion-aware ability of our proposed IML and DRC modules, we give the visualization comparison of different trajectory filling strategies using SDP detector in Fig. 5. Just like the colors used in Fig. 2 of our main work, the green boxes indicate the ground truth boxes, the blue boxes represent the inertia IML predictions, and the yellow boxes are the filled tracking fragment by our DRC module. Frame 232 corresponds to the deactivated point **A** in Fig. 2 of our main work, and frame 254 means the reconnection point **B**.

The visualization results in Fig. 5 support the discussions as in Fig. 2 of our main work vividly and consistently, that the filled trajectory using our DRC module is smoother and closer to the true trajectory with smaller offset. In contrary, if the pedestrian changes the walking direction slightly during occlusion similar to the target lady in Fig. 5, the inertia IML predictions then become further and further away from the ground truth boxes with the tracking goes on from frame 232 to frame 252, and may produce a saltation when get reconnected like the non-smooth transition from frame 252 to frame 254.

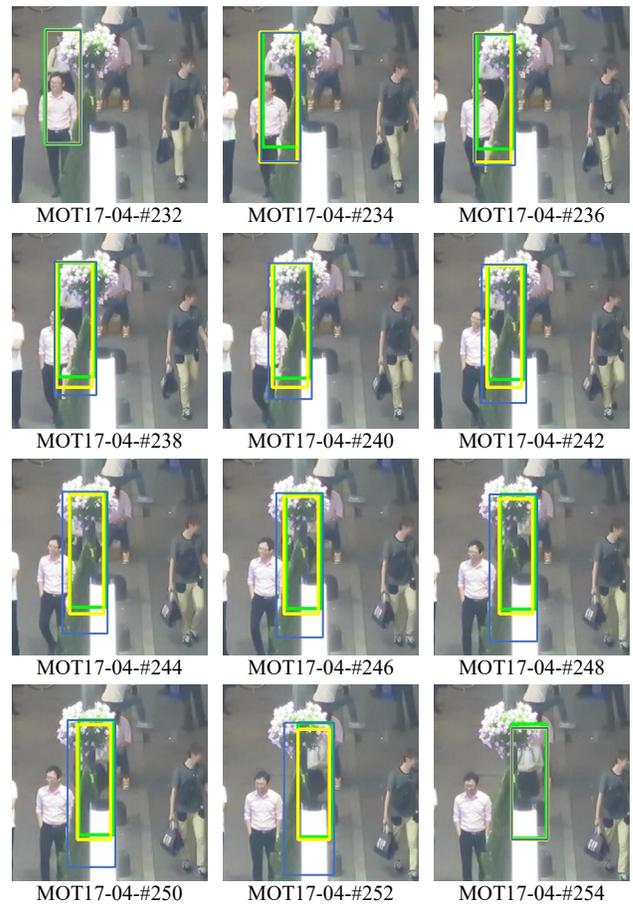

Fig. 5. Visualization comparison of trajectory filling by inertia IML (blue) or our DRC module (yellow) using SDP detector. The green boxes indicate the ground truth boxes.